%% file: main.tex
\documentclass{article}
\usepackage{ijcai21}

\usepackage{times}
\usepackage{soul}
\usepackage{url}
\usepackage[hidelinks]{hyperref}
\usepackage[utf8]{inputenc}
\usepackage[small]{caption}
\usepackage{graphicx}
\usepackage{amsmath}
\usepackage{amsthm}
\usepackage{booktabs}
\usepackage{multirow}

\usepackage{bigstrut}
\usepackage{tikz}
\def\checkmark{\tikz\fill[scale=0.4](0,.35) -- (.25,0) -- (1,.7) -- (.25,.15) -- cycle;}
\usepackage{algorithm}
\usepackage{algorithmic}

\title{Deep Learning for Virus-Spreading Forecasting: a Brief Survey}

\author{
Federico Baldo$^1$\and%\footnote{Contact Author}
Lorenzo Dall'Olio$^1$\and
Mattia Ceccarelli$^1$\and
Riccardo Scheda$^1$\and \\
Michele Lombardi$^{1, 2}$\and
Andrea Borghesi$^{1, 2}$\and
Stefano Diciotti$^{1, 2}$\And
Michela Milano$^{1, 2}$\\
\affiliations
$^1$University of Bologna \\
$^2$Alma Mater Research Institute for Human-Centered Artificial Intelligence\\
\emails
\{federico.baldo2, lorenzo.dallolio4, riccardo.scheda2, mattia.ceccarelli5, michele.lombardi2, andrea.borghesi3, stefano.diciotti, michela.milano\}@unibo.it}

\begin{document}

\maketitle

\begin{abstract}

The advent of the coronavirus pandemic has sparked the interest in predictive models capable of forecasting virus-spreading, especially for boosting and supporting decision-making processes. In this paper, we will outline the main Deep Learning approaches aimed at predicting the spreading of a disease in space and time.
The aim is to show the emerging trends in this area of research and provide a general perspective on the possible strategies to approach this problem. In doing so, we will mainly focus on two macro-categories: \textit{classical Deep Learning} approaches and \textit{Hybrid models}.
Finally, we will discuss the main advantages and disadvantages of different models, and underline the most promising development directions to improve these approaches.
  
\end{abstract}

\section{Introduction}

\input{introduction}

\section{Epidemiological Data}
\label{sec:data}

\input{data}

\section{Deep Learning for Epidemics}
\label{sec:pure_deep}

In the scientific literature, we can distinguish two main trends concerning the application of DL to virus-spreading forecasting. In particular, on the one hand, many researchers have applied the traditional and well-established DL approaches, along with some fine-tuning of the hyperparameters (we will refer to these techniques as \textit{vanilla}). On the other hand, we observe an increasingly successful use of composite methods integrating multiple types of ANN modeling different aspects of the epidemic and thus taking indirect advantage of known mechanisms of the physical phenomenon.

\subsection{\textit{Vanilla} Deep Learning}
\label{vanil}
\input{vanilla}

\subsection{Composed Deep Learning Models}
\label{composed}
\input{ml}

\section{Hybrid Deep Learning for Epidemics}
\label{sec:hyb_deep}

The widespread use of DL in different fields has led to a new breed of models:  specifically, we refer here to a class of methods that tries to improve the interpretability and accuracy of the models by adopting domain-specific techniques to guide the learning process. This phenomenon has concerned also models aimed at forecasting epidemics evolution.

\subsection{Autoregressive Ensemble}
\label{AR}
\input{autoregressive}

\subsection{Compartmental Models}
\label{causal}
\input{causal}

\newcommand\Tstrut{\rule{0pt}{2.6ex}}         % = `top' strut
\newcommand\Bstrut{\rule[-0.9ex]{0pt}{0pt}}   % = `bottom' strut

\begin{table*}[ht!]
\tiny\sf\centering
%\hspace*{-1cm}
\begin{tabular}{|c | c | c | c | c | c | c | c | c | c | c |}
%\begin{tabular}{l  c c  c c c c c c c c c c c}
% \hline

\hline\Tstrut
Category & Sub-Category & Paper & Type & S & E & C & OD & MR & DA & X\\
  
\hline\Tstrut

\textbf{Deep Learning}       & \hyperref[vanil]{Simple}  & \cite{Sahid2020}  &  \hyperref[vanil]{Bi-LSTM} & & & & \checkmark & \checkmark & &\\ \cline{3-11}\Tstrut 
                             &         & \cite{8581423} & \hyperref[vanil]{LSTM} & \checkmark & & & \checkmark& \checkmark &   & \checkmark \\ \cline{3-11}\Tstrut
                             &         & \cite{Kunjir:Comparative} & \hyperref[vanil]{CNN} & & & & \checkmark & \checkmark & & \\ \cline{3-11}\Tstrut
                             &         &\cite{Shastri2020}  & \hyperref[composed]{Conv-LSTM} &  &   & & \checkmark &\checkmark & & \\ \cline{3-11}\Tstrut
                             &         & \cite{Huang:Multiple} & \hyperref[vanil]{CNN} & & & & \checkmark & \checkmark & &  \\\cline{2-11}\Tstrut
                             % &         & \cite{tcn} & TCN & 
                             & \hyperref[composed]{Composed} & \cite{10.1145/3292500.3330917} & \hyperref[composed]{AE + LSTM + clustering}  & & & \checkmark & \checkmark & & & \\\cline{3-11}\Tstrut
                             % &         &\cite{Zhu2019}  & Att-MCLSTM &  &  &  & &\checkmark & & \checkmark\\\cline{3-11}\Tstrut
                             &         &\cite{Shastri2020}  & \hyperref[composed]{Conv-LSTM} &  &   & & \checkmark &\checkmark & & \\ \cline{3-11}\Tstrut
                             &         &\cite{Ibrahim2020} &  \hyperref[composed]{VAE + LSTM} & \checkmark &   & & & \checkmark & & \checkmark\\ \cline{3-11}\Tstrut
                             &          &  \cite{Huang:Novel} & \hyperref[composed]{CNN + Bi-GRU} &  \checkmark & & & \checkmark & \checkmark & & \\ \cline{3-11}\Tstrut
                             &          &  \cite{Wu:DeepLearning}   & \hyperref[composed]{CNN + GRU} & \checkmark &  & \checkmark & \checkmark & \checkmark & & \\ \cline{3-11}\Tstrut
                             &          & \cite{10.1007/978-3-030-65347-7_35} & \hyperref[composed]{GNN} & & & \checkmark & & & & \\ \cline{3-11}\Tstrut
                             &          &  \cite{10.1145/3340531.3411975} & \hyperref[composed]    {GNN + RNN + CNN} & \checkmark   &  & & \checkmark & \checkmark & & \\ % \cline{3-11}\Tstrut \\
                             % &          &  \cite{Li:GSRNN} & GSRNN  &  \checkmark  &  & & \checkmark &\checkmark &  & \\
\hline\Tstrut
\textbf{Hybrid Deep learning}    & \hyperref[AR]{Autoregressive} & \cite{CHAKRABORTY2019121266} & \hyperref[AR]{ARIMA + NNAR} & & & & \checkmark & \checkmark & & \\\cline{3-11}\Tstrut
                                  & & \cite{10.1371/journal.pone.0156768} & \hyperref[AR]{ARIMA + NNAR} &  &  & & \checkmark & & & \\\cline{3-11}\Tstrut
                                   & & \cite{wang2020advanced} & \hyperref[AR]{SARIMA + NNAR} &  &  & & \checkmark& & &  \\\cline{3-11}\Tstrut
                                   & & \cite{Wange024409} & \hyperref[AR]{SARIMA + NNARX}   &  &  & & \checkmark & & & \checkmark \\\cline{2-11}\Tstrut

                                & \hyperref[causal]{Compartmental}     &  \cite{FAROOQ2020110148} & \hyperref[causal]{FFNN + SIRVD} &  & \checkmark & & & & & \\\cline{3-11}\Tstrut
                                & & \cite{Jo2020.04.13.20063412} & \hyperref[causal]{FFNN + SIR} &  & \checkmark & & & \checkmark & & \\\cline{3-11}\Tstrut
                                &   & \cite{TDEFSI} &\hyperref[causal] {LSTM + SEIR}  &  & & & & \checkmark & \checkmark & \\
\hline
\end{tabular}
\caption{Features of the models: Spatial Data (S): spatial or geographical data
; Explainabile (E); Code available (C); Open Data (OD): data are publicly available; Multiple Countries/Regions (MR): the model considers multiple regions and countries; Data Augmentation (DA); Contextual/exogenous information (X): the model is integrated with input not directly related to the learning task, but that affect the outcome of the prediction.}
\label{tab:booktabs2}
\end{table*}
%\section{Related Areas}

\section{Discussion}
\label{sec:disc}

\input{discussion}

\section{Conclusion}
\label{sec:conclusion}

\input{conclusion}

\end{document}

%% file: introduction.tex
The 21st century has been marked by an ever-increasing global interconnection, which, on the one hand, created new possibilities for human activities and technological improvement, while, on the other hand, has paved the way to the rapid spreading of new pathogens. Many scientific reports are linking environment-invasive human activity to the emergence of novel zoonotic viruses \cite{bengis2004role}. Among these, \textit{airborne} diseases are the most worrying, due to their capability of spreading at a fast pace (e.g, SARS, MERS, H1N1).
% ORIGINALLY: through the air
Most notably, the recent outbreak of Sars-CoV-2 \cite{velavan2020covid}, which started in Wuhan, China, in 2020 (possibly even 2019), has been a great challenge for our society.

For this reason, the presence of predictive systems capable of forecasting the dynamics of the virus on a short and long-term range is of crucial relevance to counteract the surging of an epidemic. The decision process regarding policies apt at containing the spreading of the virus is dependent on accurate predictive systems since they allow the evaluation of different scenarios and intervention plans.
To this end, many epidemiological models have been developed in the past to approximate the dynamics of a disease \cite{duan2015mathematical}. However, in light of the recent increase in computational capability, new possibilities are emerging.

In the last years, we are witnessing a booming application of Machine Learning (ML) to create predictive models. The high volume of information available through Information and Communication Technologies, particularly smartphones and social media, has driven the research towards the use of ML-based predictive models.
Among these, Deep Learning (DL) has proved to be one of the most effective tools, thanks to its capability to adapt to a wide variety of problems.

In this survey, \textit{we focus on the application of DL models to approximate the dynamics of diseases and to produce forecasting systems capable of predicting the spreading of an airborne virus}. We decided to keep the scope of the considered illnesses as wide as possible.
However, it is impossible not to notice the wealth of many recent works regarding COVID-19. Nevertheless, the approaches outlined in this paper can be applied to any disease if proper modifications are made.

One of our goals is to provide a general taxonomy of the different DL methodologies that can be deployed to tackle this problem. 
The related scientific literature follows two main directions: one is focused on the application of \textit{pure} DL models, in either a \textit{vanilla} fashion or with a combination of different Artificial Neural Networks (ANN),  in an attempt to make predictions with minimal assumptions or human intervention; the other one tries to incorporate well-established epidemiological methods with DL, to take advantage of decades worth of models and insights, and produce more explainable predictions.
In particular, the second class of methods relies on the idea of embedding domain-specific knowledge into the predictive model, a new trend in the field of ML. The principle at the foundation of this approach is to integrate into the learning process prior information, or techniques, relative to the field of interest to improve performances and build custom models (e.g., \textit{end-to-end learning}).

The rest of the paper is structured as follows: in Section~\ref{sec:data}, we briefly discuss epidemiological data; we will then proceed with a comparative revision of the different approaches: \textit{Deep Learning for Epidemics} (Sec.~\ref{sec:pure_deep}) and \textit{Hybrid Deep Learning for Epidemics} (Sec.~\ref{sec:hyb_deep}); next, in Sec.~\ref{sec:disc}, we will discuss and compare the approaches presented in the previous sections; Sec.~\ref{sec:conclusion} concludes the paper.

%% file: data.tex
Deep Learning is a data-driven approach, therefore its performance relies on the presence of accurate and consistent data. For this reason, the data-gathering strategy is of high relevance when approaching a learning task. In this view, the coronavirus pandemic has allowed collecting a large amount of information, being the first global epidemic in an era marked by the presence of smartphones and big data systems.

We can distinguish two categories of data regarding a virus, which implicitly define different learning objectives: \textit{clinical data}, describing medical aspects of the disease (e.g., hospital records, symptoms, administration of drugs), and \textit{spatio-temporal data}, underlying the spreading of the virus, usually in terms of infected, recovered and deceased individuals. Many scientific reviews have historically focused on the first category \cite{8990083}, whereas in this paper we will only focus on the second one, examining techniques approximating the dynamics of the virus and forecasting its spreading.

Besides data strictly concerning the disease, information describing contextual aspects of the epidemic might help to improve the accuracy of the predictive model. For instance, knowledge on the trend of temperature in specific regions might increase the chances of capturing a seasonal behavior of the virus, if present, whereas information regarding population density allows to identify scenarios with high contact rates among individuals, hence increasing the likelihood of contagion. Moreover, for most epidemics, countries cannot be considered isolated due to the highly infectious nature of some viruses and the frequent population flows between borders. Consequently, the information regarding neighboring countries could consistently improve the understanding of the epidemic.

We shall also consider which technologies could be used to acquire data regarding the spreading of the virus. Among the main sources of information we can distinguish:
\begin{itemize}

\item \textbf{Government and health records}: governments, hospital administrations, and health organizations can voluntarily release records regarding citizens, either publicly, or in the context of a specific project. For instance, the Centers for Disease Control (CDC) have published their records on seasonal flu \cite{10.1145/3292500.3330917}.  

\item \textbf{Mobile data}: the increasing presence of smartphones in our lives has made it easier to gather information about our habits, individual movements, and contact patterns \cite{oliver2020mobile}, producing valuable information for both epidemic control and virus-spreading forecasting.  

\item \textbf{Social media and search engines}: along with the growing use of smartphones, social media and search engines can be great sources of information, ranging from reported symptoms, inferred from public posts, to search queries. One interesting example is shown in \cite{santos2014analysing}, where Twitter messages and web queries were used to estimate the flu trend.

\item \textbf{Sensors}: information can also be collected through the use of specific pieces of hardware, or sensors, as shown in \cite{Sareen2018IoTbasedCF}, where Radio-Frequency-Identification (RFID) technology is used to collect data regarding the contacts among individuals in a small environment, and then recreate a possible chain of infections.
\end{itemize}

The presence of a data-gathering infrastructure has improved the quantity and quality of data available to the research. However, there are many issues concerning the possible consequences on human privacy and personal freedom, with an increasing demand for stricter rules and a secure design of these systems \cite{simko2020covid19}.

Ultimately, the quality of DL models depends on the amount of data available, but, unfortunately, information can be scarce, especially during the first stages of the epidemic. Some authors have specifically addressed this issue with data augmentation methodologies, as proposed in \cite{TDEFSI}, or smart strategies to exploit historical data \cite{10.1145/3292500.3330917}.

%% file: vanilla.tex
In this section, we revise the application of traditional DL methodologies to predictive epidemiology. Here we observe the prevalence of two perspectives: on one side, the application of techniques historically designed to model sequential data, therefore particularly suited in presence of temporal data (i.e., time series), while on the other, the implementation of DL approaches for spatial information transformed to handle temporal information. 

\paragraph{Recursive Neural Networks.} \label{Recursive Neural Networks}
Traditionally, this class of methods has been widely applied to approximate the dynamics of epidemics (and more in general sequential data), thanks to their structure and capability of developing a \textit{long-term memory}. In the RNNs family, we include Long-Short Term Memory Networks (LSTM) and Gated Recursive Unit (GRU), which are particularly suited for temporal series. The standard application of RNNs involves the construction of an appropriate dataset by arranging each feature according to the number of \textit{look-back days}, namely, how many days in the past are taken as input when making a prediction. This allows incorporating in the forecasting process information regarding the general trend of the epidemiological curve.

A recent application of RNNs can be seen in \cite{Sahid2020}, where authors have used LSTM and GRU to model the COVID-19 pandemic data (i.e., number of infected, deceased, and recovered cases) of different countries, with results that show how these methods outperform more classical ML approaches in terms of prediction accuracy on historical data, such as Support Vector Regression (SVR), Radial Basis Function (RBF) kernels and Auto-Regressive Integrated Moving Average (ARIMA).  
In the same study, the authors improved the model using Bidirectional LSTM (Bi-LSTM) to enhance the capability of developing a long-term memory in the model. While classical RNNs-based approaches process sequential data in a specific order, i.e., from past to future, Bi-LSTM allows the information to flow in both directions of the time series (i.e., past-future, future-past), with an increase in performance, also w.r.t. LSTM and GRU.

In \cite{Shastri2020} is proposed a method to bring further the application of LSTM on epidemiological data. The idea is to stack multiple LSTM layers, where each intermediate layer output is used as input for the next one, resulting in more complex and deeper models. This method, called Stacked-LSTM, seems to increase significantly performance due to its higher capability to abstract the input through the stratification of the layers.

While Recursive networks produce an accurate approximation of time series, they often do not include contextual information, such as spatial or seasonal data, that might be relevant to consistently capture the dynamic of a virus. This issue is directly tackled in \cite{8581423}, where authors proposed an LSTM-based method incorporating geographical proximity information and climate variables (e.g., humidity, temperature, precipitation) to predict the number of flu cases in the population. The approach consists of two steps: in the first stage, the LSTM neural network is trained on the flu time series, while in the second stage, the impact of climatic variables and spatio-temporal adjustment is added to the flu counts estimated by the LSTM model. The \textit{contextual features} are integrated into the learning process via a linear combination of the variables and multiplied with the actual predicted number of daily new cases.

\paragraph{Convolutional Neural Networks.} This DL tool is one of the most widely used approaches to deal with spatial proximity within input data. 
Such aspects can be very useful in forecasting epidemiological trends. As CNNs have been proved useful at dealing with spatial proximity (by extracting features embedding the spatial locality), they can be adopted for dealing with temporal locality as well, as in the case of ordered sequential data.
 
Most basic CNNs applications rely on the use of ordered time series as input and require the model to predict the next data points in the series; however, CNNs are known to be prone to overfitting historical data. Nevertheless, this can be sufficient in presence of poor contextual data to outperform LSTM or simpler algorithms such as decision trees (DT) \cite{Kunjir:Comparative}. 
Predictions can be improved by feeding the models with multiple correlated time series. 
For example, in \cite{Huang:Multiple} authors use the data relative to confirmed, deceased, and recovered cases to predict the daily count of infected. The CNN is compared with LSTM, GRU, and MLP models on 7 different Chinese cities showing smaller accuracy error w.r.t. all other models, especially MLP. 

A method that exploits CNNs, and that has not been sufficiently explored yet in the field of predictive epidemiology, is Temporal CNN (TCN), which showed to have comparable, if not better, performance w.r.t. RNNs. The approach was first presented in \cite{tcn} and it is based on the deployment of \textit{causal} convolution for \textit{sequence-to-sequence} learning tasks. In this method, the kernel is applied only over data relative to the present time step or previous input points in the sequence, avoiding information leakage from future to past. The network architecture is based on 1-D fully convolutional layers with the same length as the input data.
This methodology is gaining increasing attention from the DL community and we regard it as a valuable approach going forward in the research of forecasting systems for epidemics.

%% file: ml.tex
In many instances, combining multiple DL-based methodologies can greatly improve the accuracy of the final predictions. For instance, integrating two models, one approximating the temporal features and the other encapsulating spatial characteristics, can increase the overall performance. For this reason, a plethora of works integrating one or more DL approaches has been proposed in recent years.

\paragraph{Recursive Networks Autoencoders.}

An interesting set of DL-based methods, applied in the field of predictive epidemiology, deploy the so-called LSTM Autoencoders.
In general, autoencoders are composed of two parts: an \textit{encoder}, which creates an internal representation of the input sequences, namely the internal state of the LSTM cells used to model the sequence, and a \textit{decoder}, which maps the embedding into the output feature space (i.e., a series of dense layer minimizing a reconstruction loss). 

An application of this method is proposed in \cite{10.1145/3292500.3330917}, where it has been deployed in the context of the FluSight Task\footnote{\url{https://www.citizenscience.gov/catalog/171/}}, a challenge launched by the CDC to advance the research w.r.t. seasonal influenza forecasting. The proposed model is trained over the historical data relative to seasonal flu in the last 20 years. The authors implemented a system based on an LSTM paired with an attention mechanism that embeds the historical data into a latent space, whereas the decoder performs the actual predictions. The data relative to the trend of the current flu season is matched to the most similar historical series embedding through a Deep Clustering technique on the latent space. The historical epidemiological curve is reconstructed and used to predict the evolution of the ongoing outbreak of flu.

%In \cite{Zhu2019}, such method is applied to Influenza-Like Illness (ILI) forecasting for the Guangzhou region in China. The authors developed a multi-channel LSTM neural network (or MCLSTM), adding an attention mechanism (Att-MCLSTM) to improve the prediction accuracy. The model takes into account data regarding the seasonal flu and context features relative to the weather (e.g., temperature, humidity). The model is then divided into two LSTM branches, one approximating the flu data and the other the climate-related information. The output of the two channels is eventually merged and passed to the attention layer. The attention mechanism is inserted to selectively learn the input streams by conserving the intermediate outputs of LSTM.

Another example, inspired by Variation Autoencoder, was proposed by \cite{Ibrahim2020}. The model, which is based on a variational LSTM-Autoencoder is composed of two branches trained in parallel. The first branch is a self-attention LSTM encoder, whose inputs are the daily new infected cases, along with government policies and \textit{urban features} (population density, fertility rate, etc.).
The second branch is an encoder, whose input is a distance-weighted adjacency matrix, computed using the geographical locations of the countries. The output of the encoder is a set of latent variables that are concatenated with the output of the first branch and passed to an LSTM which outputs the prediction of the daily cases per country. 
%The idea behind this model does not rely only on sequential data regarding the virus, but also urban characteristics represented by locational and demographic data, and intervention policies directed at containing the outbreak.

\paragraph{Combining Recursive and Convolutional Neural Networks.}

We have seen in the previous section how both RNNs and CNNs can be suitable choices to model the dynamics of a disease. An application combining both of these approaches is proposed in \cite{Shastri2020} with Convolutional LSTM (Conv-LSTM). The model consists of a Stacked-LSTM, where instead of traditional matrix multiplication, a convolutional operator is alternated for each layer. 
The model is compared to other RNNs methods, included Bi-LSTM and Stacked-LSTM, showing improved accuracy.

On the same line, a method combining temporal information, i.e. 1D time series, with spatial information such as the disease dynamics of neighboring countries is introduced in \cite{Huang:Novel}. The model combines the results of two different pipelines: the first considers as input the number of new daily recovered cases, deaths, confirmed cases, and the cumulative count of these features of the past 5 days. 
Such input is given to a 1D-CNN serially combined with a bi-GRU to extract temporal patterns from the input trends. 
The second pipeline, given the 2D matrix representing the epidemic trend of different regions in time, extracts the spatial correlation patterns through the use of a 2D-CNN. 
Such input matrix models the presence in the epidemic trends of highly correlated and dependent neighboring countries. 
Ultimately, the outputs of both pipelines are concatenated, and a dropout layer is used to prevent overfitting in case of scarce data. 
 
A similar method was proposed by \cite{Wu:DeepLearning} for the influenza epidemic.
The researchers considered the data of the USA and Japan, i.e., the weekly reported patients count over a single state or prefecture.
In this work, the authors used CNNs to fuse information from different data sources (e.g., different regions) and GRUs to capture the long-term correlation in the time series, with a residual structure. The residual structure is adopted to introduce highly relevant long-jumps in the information flow (e.g., capture annual epidemiology patterns).
We can observe that the CNNs were used to learn correlations \textit{among} signals, while GRUs were targeting correlation \textit{within} every single signal. 
However, the truly innovative solution provided in this work is the use of an adjacent nearest neighbor matrix as a filter of the convolutional kernel of each country. 
This sort of adjacent convolution gets parameters focused on more local information in opposition to the classical grid convolution, implying that a single filter could be able to represent more complex patterns, which are usually captured by multiple filters when using the classical grid convolution. 

\paragraph{Graph Neural Networks}

Geometric Deep Learning, and more specifically Graph Neural Networks (GNNs), has produced great results in many fields of research. Even if the use of GNNs is currently limited in computational epidemiology, we consider this as one of the most promising frameworks for virus spreading forecasting. Many of the traditional epidemiological models have deployed complex networks to describe the dynamic of epidemics. Consequently, GNNs are capable of extending the principle at the foundation to such data structure. For this reason, we reckon that a very promising direction consists in exploring GNNs which explicitly model the graph-based structure encoding the topological relations at the heart of epidemics spreading (e.g., interaction among people, geographical distributions, etc).

A variety of GNNs applications to the problem was studied in  \cite{10.1007/978-3-030-65347-7_35}, where multiple declinations of this approach were tested on a synthetic dataset. The main idea is to model the spreading starting from a network of individuals (i.e., the nodes) that can interact in specific ways (i.e., the edges). Then approximate the function mapping the vertex into a positive real number representing the count of infected nodes through \textit{node regression}, assuming to have some initial contagions in the population. This method can be either based on pure GNNs or on propositional learners, where the first uses the adjacent matrix during the training of the model while the second does not. The node regression was ultimately implemented with either graph-based methods, such as \textit{Graph Attention Networks} and \textit{Graph Isomorphic Networks}, or a Gradient Boosting model, namely XGBoost. 

Another example of the application of this class of methods is provided in \cite{10.1145/3340531.3411975} with Cola-GNN, a model aimed at producing reliable long-term epidemiological predictions. The Cola-GNN is one of the first original works to apply Graph Neural Networks in epidemic forecasting. The framework is composed of three main components: a \textit{Location-aware Attention} element for capturing spatial local dependencies through an RNN, a \textit{Temporal Convolution Layer}, in charge of approximating the temporal features, and a \textit{Global Graph Neural Network}, that encapsulates the outputs of both previous layers and updates the nodes in a \textit{message-passing} fashion, modeling the spreading of the virus.

%% file: autoregressive.tex
Among the most widely used methods to approximate time series, there are Auto-Regressive models (AR). This class of models offers a variety of tailored solutions capturing different aspects of sequential data, such as Vector Auto-Regressive models (VAR), Auto-Regressive Integrated Moving Average (ARIMA), and Seasonal ARIMA (SARIMA). The key idea behind AR is to forecast the future using past information, more specifically a linear combination of the predictors. While these methodologies have proved to be effective in the case of epidemiological data, often the nature of the underlying AR technique is not suited to capture nonlinearities in the features. Many studies are tackling directly this issue, mainly combining both AR and DL to mitigate their respective deficiencies in an ensemble fashion.
In \cite{CHAKRABORTY2019121266} and \cite{10.1371/journal.pone.0156768} the focus is precisely on approximating both linear and non-linear aspects of the data; this is achieved by combining an ARIMA model and a Neural Autoregressive Network (NNAR) on the residuals.   \cite{wang2020advanced} propose a slightly improved version of this approach, relying on the SARIMA model, which explicitly encapsulates the seasonal component
The fusion of AR and DL shows a steady improvement in the accuracy of the prediction w.r.t. \textit{vanilla} approaches. However, these models consider only the temporal component of the data, while the usage of contextual information and spatial data could be decisive to obtain high-quality forecasts. 
A step in this direction is proposed in \cite{Wange024409}, where the aforementioned method based on SARIMA, is improved by integrating into the training process \textit{exogenous inputs}, namely a set of variables or a series that are not directly related to learning target, but that influence it.

%% file: causal.tex
A recent trend in the development of ANN-based methods relies on the integration of domain-specific knowledge \cite{8621955}. The general idea is to design the learning process to incorporate data and techniques that are specifically referred to some particular aspect of the domain in consideration. In the case of predictive epidemiology, this tendency has materialized in hybrid approaches integrating compartmental models and deep learning. Historically, many forecasting methods for virus-spreading have been proposed \cite{duan2015mathematical}. Among the most notable we find compartmental models, i.e. mathematical models where the population is divided into categories representing different conditions w.r.t. the epidemic, and that describe the progression of the virus through differential equations (e.g. SIR: \textbf{S}usceptible-\textbf{I}nfected-\textbf{R}ecovered). 

In \cite{FAROOQ2020110148} this idea is applied to predict the evolution of the COVID-19 epidemic in India. More specifically, the proposed framework is based on a SIRVD (\textbf{S}usceptible-\textbf{I}nfected-\textbf{R}ecovered-\textbf{V}accinated-\textbf{D}eceased) dynamic system which approximates the epidemiological curve; the parameters describing the transition into the different compartments are learned through an incremental learning approach. The idea is that a Feed-Forward Neural Network (FFNN) can be used to iteratively approximate the incoming data, which is updated based on the progression of the pandemic; this allows to improve the model without training the ANN on the whole dataset every time is updated. In similar fashion \cite{Jo2020.04.13.20063412} proposed a \textit{forward-inverse} neural network for SIR models. This solution can be considered as a sort of \textit{end-to-end} approach where the parameters and the compartment (or category, such as the infected individuals in the population) at each time step are estimated with a time-dependent FFNN based on the historical data of COVID-19 in South Korea. This model shows the gain in terms of explainability deriving from the use of a hybrid framework.

Another model exploiting compartmental systems was introduced by \cite{TDEFSI}. In this case, the proposed methodology attempts to solve problems related to scarce data with a data augmentation framework for forecasting the incidence of Influenza-Like Illnesses (ILI) in the USA. The model, which is called Theory Guided Deep Learning Based Epidemic Forecasting with Synthetic Information, or TDEFSI, integrates the strengths of deep neural networks and high-resolution simulations of epidemic processes over complex networks.
The data augmentation is performed by generating synthetic data using computer-based simulations of causal processes that capture epidemic dynamics: for instance, given a US state and an existing disease model (e.g., \textbf{S}usceptible-\textbf{E}xposed-\textbf{I}nfected-\textbf{R}ecovered), a marginal distribution is estimated for each parameter in the model. Then, a synthetic training dataset is generated for the target US state by sampling from the learned marginal distributions.
The core model is composed of a two-branch LSTM-based deep neural network that captures the temporal dynamics of both within-season observations and between-seasons observations.
The main advantage of this method is the capability to synthesize a large volume of time series from simulations based on epidemiology theory, reducing the risk of overfitting while training the model. At the same time, this method introduces the challenge of minimizing the gap between synthetic and real data, which is a non-trivial problem in itself.

%% file: discussion.tex
In the previous sections, we have outlined two trends in the creation of epidemiological DL-based predictive models.
We observed a group of methods that adopt \textit{pure} DL methods, using either a conservative (i.e., \textit{vanilla}) application of ANNs or more sophisticated implementations combining different models to target varying aspects in the data.
The other trend is focused on hybrid DL models. These methods focus on finding ways to combine traditional techniques of predictive epidemiology, namely autoregression and compartmental dynamic systems, with ANNs.

Both approaches seem to be effective to solve the task at hand (i.e. epidemic spreading forecasting), however, we would like to point out some of the main advantages and disadvantages of these methodological strategies.

The flexibility and the abstraction provided by deep ANNs generate accurate models, which proved themselves capable of capturing at least the general trend of the epidemic. Moreover, the learning process requires minimal assumptions on the problem at hand and almost no human intervention during the training process. This can be extremely helpful in case of new and unexplored situations, such as COVID-19, where we want to analyze the phenomenon without biases. Pure Deep Learning approaches can also adapt naturally to evolving conditions, such as changes in the mobility patterns and daily routines emerging as a reaction to the pandemic.

We have also observed how these models can be improved consistently in two ways: first, by adding contextual information such as seasonal or urban-environment features, and second, developing more sophisticated approaches, which, for instance, combine different types of networks. This is especially useful when dealing with spatial and temporal data, as each sub-network can deal with a particular aspect of the data, in a way decomposing the overall problem.

Ultimately, beyond the more classical applications of RNNs and CNNs, we identified new possible approaches to build epidemiological predictive systems. In particular, we have briefly discussed the deployment of TCNs and GNNs and they could lead to an improvement of the forecasting accuracy. Unfortunately, these areas have been relatively understudied so far, and we reckon that there lies a great potential ripe to be exploited; we hope that future research will explore these directions. Above all, we believe GNNs could allow us to formulate the learning task in a way that is suited to represent epidemiological scenarios, and represent spatio-temporal data, as well as mimicking the processes underlying the spreading of a virus.

Pure DL methodologies do not provide the possibility of exploiting the domain knowledge available in the field of predictive epidemiology. Moreover, these models are lacking in terms of interpretability of the results, which might be a crucial aspect, especially if the predictions are the basis for a decision process; broadly speaking, the possibility to interpret and explain the predictions is a mandatory requirement wherever human health is concerned \cite{ahmad2018interpretable}.

To this end, hybrid models tend to be more effective at incorporating well-established practices and provide interpretable results. In particular, the integration of compartmental models and DL seems to be one of the most promising research directions to explore. These models allow to obtain explainable predictions, based on simple and known principles of epidemiology. However they might also inherit some of the assumptions and rigidity of the classical approaches (e.g., a limited amount of compartments to describe the epidemic), therefore they might require some further elaboration to be adapted to a specific disease. Fortunately, compartmental models are highly customizable \cite{giordano2020sidarthe}, therefore easy to mold on the virus under scrutiny.

Ultimately, the integration of domain-specific methods into the learning process might help generalize the model beyond the training data, proving to be quite useful in presence of scarce information on the epidemic.
For this reason, we believe that hybrid models could be more suited to support specific containment policies and provide the means to evaluate different scenarios relative to multiple intervention plans.

An important area of open research, where little progress has been currently made
%\footnote{To the best of our knowledge, one of the few approaches is unpublished and available at \url{https://github.com/leaf-ai/covid-xprize/tree/master/covid_xprize/standard_predictor}}
, involves how to best take into account the effect of containment measures on the spread of an epidemic. This ability is a key requirement to support decision-makers in the identification of the best approach for managing the trade-off between health, societal, and economic costs. Taking into account these factors in primarily data-driven methods can be very challenging, since data about containment measures is generally scarce (especially at the beginning of an epidemic), and data about historical epidemics may not be entirely reliable when a new pathogen appears.

%% file: conclusion.tex
The recent outbreak of coronavirus has proved our unpreparedness relatively to unexpected scenarios. In this context, the availability of accurate predictive models capable of forecasting the virus spreading in space and time is crucial to make informed and scientifically proved decisions to stem the virus. 
The use of DL to solve this problem has widely increased, showing interesting improvements w.r.t. more classical approaches. Given the current situation, this class of methods will be increasingly applied to tackle this problem in the coming years. For this reason, we have attempted to outline the main trends in this field, hoping to provide a starting point for readers approaching this challenge.